\documentclass[10pt,twocolumn,letterpaper]{article}

\usepackage{cvpr}
\usepackage{times}
\usepackage{epsfig}
\usepackage{graphicx}
\usepackage{amsmath}
\usepackage{amssymb}
\usepackage{slashbox,multirow}
\usepackage{setspace}
\usepackage[pagebackref=true,breaklinks=true,letterpaper=true,colorlinks,bookmarks=false]{hyperref}

\newcommand*{\affaddr}[1]{#1} % No op here. Customize it for different styles.

\newcommand*{\email}[1]{\texttt{#1}}

\cvprfinalcopy % *** Uncomment this line for the final submission

\begin{document}

%%%%%%%%% TITLE
\title{DivCo: Diverse Conditional Image Synthesis via\\ Contrastive Generative Adversarial Network}

\author{%
Rui Liu \quad Yixiao Ge \quad Ching Lam Choi \quad Xiaogang Wang \quad Hongsheng Li \\
\affaddr{Multimedia Laboratory, Chinese University of Hong Kong}\\
\email{\{ruiliu@link, yxge@link, hsli@ee\}.cuhk.edu.hk}\\
}

\maketitle
%\thispagestyle{empty}

%%%%%%%%% ABSTRACT
\begin{abstract}
    Conditional generative adversarial networks (cGANs) target at synthesizing diverse images given the input conditions and latent codes, but unfortunately, they usually suffer from the issue of mode collapse.
    To solve this issue, previous works \cite{bicyclegan,DRIT} mainly focused on encouraging the correlation between the latent codes and their generated images, while ignoring the relations between images generated from various latent codes.
    The recent MSGAN \cite{msgan} tried to encourage the diversity of the generated image but only considers ``negative'' relations between the image pairs.

    In this paper, we propose a novel DivCo framework to properly constrain both ``positive'' and ``negative'' relations between the generated images specified in the latent space. To the best of our knowledge, this is the first attempt to use contrastive learning for diverse conditional image synthesis.
    A novel latent-augmented contrastive loss is introduced, which encourages images generated from adjacent latent codes to be similar and those generated from distinct latent codes to be dissimilar. The proposed latent-augmented contrastive loss is well compatible with various cGAN architectures.
    Extensive experiments demonstrate that the proposed DivCo can produce more diverse images than state-of-the-art methods without sacrificing visual quality in multiple unpaired and paired image generation tasks.
   Training code and pretrained models are available at \url{https://github.com/ruiliu-ai/DivCo}.
\vspace{-2em}
\end{abstract}

%%%%%%%%% BODY TEXT
\section{Introduction}

Generative adversarial network (GAN)~\cite{gan} has shown great potential to capture complex distributions and generate high-dimensional samples since its first introduction in 2014. The follow-up years witnessed its great progresses and successes in synthesizing realistic high-resolution images~\cite{dcgan, lsgan, wgan, proggan, stylegan, biggan}. Based on GANs, conditional generative adversarial networks (cGANs)~\cite{cgan} were proposed, which focus not only on producing realistic images but more on preserving the input conditional information.
For example, ACGANs~\cite{acgan} generated diverse images conditioned on class labels. Pix2pix~\cite{pix2pix} and CycleGAN~\cite{cyclegan} translated images across two characteristic domains to change their visual styles under paired and unpaired settings, respectively.
There exist other cGANs that condition on input images, \textit{e.g.} super-resolution~\cite{srgan}, style transfer~\cite{UNIT, DRIT}, inpainting~\cite{inpainting-ca}, denoising~\cite{denoisinggan}, \textit{etc.}, as well as cGANs conditioned on text descriptions~\cite{stackgan, stackgan2, attngan}.

\begin{figure}[t]
    \centering
    \includegraphics[width=1\linewidth]{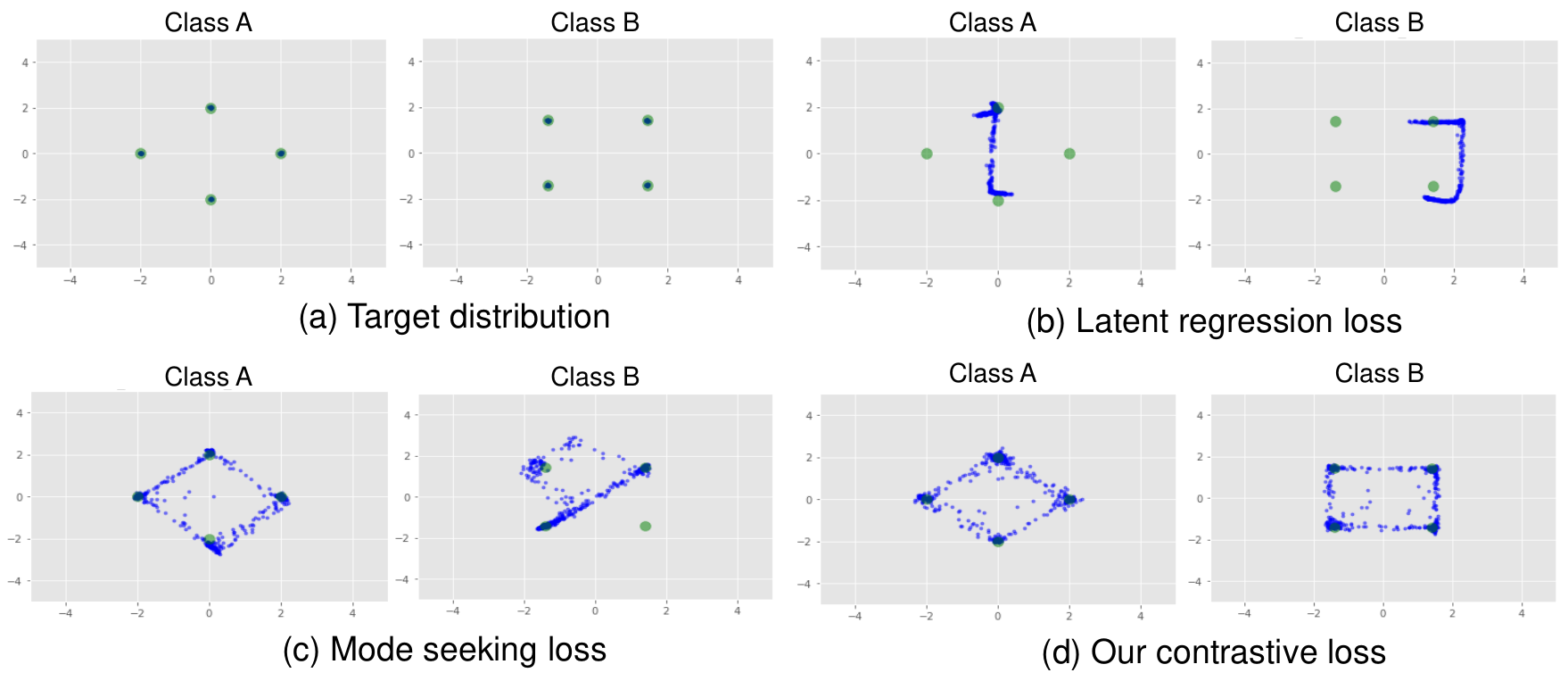}
    \caption{2D toy experiment for demonstrating the generation distribution learned with different regularization terms. (a) The ground-truth distributions for the two different classes, each of which is a Gaussian mixture model. (b) The generated samples from the learned distribution by the latent regression loss \cite{bicyclegan} shows its weakness in covering all modes of the ground truths. (c) The generated samples from the learned distribution by the mode seeking loss \cite{msgan} shows similar patterns regardless of different classes. (d) The generated samples from the learned distribution by our proposed latent-augmented contrastive loss demonstrates an unbiased distribution which is properly dependent on both conditional input and latent codes.}
    \label{fig:toy}
\vspace{-1em}
\end{figure}

Real-world scenarios expect the synthetic samples to be diverse and able to manipulate flexibly. However, all the applications mentioned above suffer from the problem of mode collapse to some extent, even though randomly sampled latent codes are added as additional inputs.
To deal with this shortcoming, many works attempted to enhance the correlation between input latent codes and output images to ensure that the latent codes have control over the generated images.
BicycleGAN~\cite{bicyclegan} and DRIT~\cite{DRIT} adopted a latent regression loss term, which encourages the model to recover the input latent code from the generated images.
However, the effect of this term on boosting the generation diversity is far from satisfactory (see Fig.~\ref{fig:toy}(b)), which is due to the fact that this term only considers the relation between individual latent codes and their generated images.
The relations between images generated from various latent codes are more valuable but were neglected.
MSGAN~\cite{msgan} tried to tackle the problem and proposed a mode seeking loss, which aims to improve generation diversity by maximizing the dissimilarity of two arbitrary images. However, the distance between two sampled latent codes may be close to each other and their synthesized images should not be pushed away.
By imposing such strong yet sub-optimal constraints between pairs of generated images, the learned distribution easily turns out to be biased, \textit{i.e.} the generation
results only depend on latent codes while ignoring the conditional input (see Fig.~\ref{fig:toy}(c)).
Such a phenomenon can also be observed in our later experiments (refer to Fig.~\ref{fig:paired}(b) and Fig.~\ref{fig:dog2cat}).

We argue that the unsatisfaction of MSGAN~\cite{msgan} derives from its strategy on always treating any image pairs as ``negative'' pairs while ignoring ``positive'' pairs.
However, they should be equally crucial for the generator to correctly capture the semantics of various latent codes.
Towards this end, we attempt to learn unbiased distributions by considering the ``positive'' and ``negative'' relations simultaneously in the form of contrastive learning.
Contrastive learning was widely applied in self-supervised representation learning tasks \cite{precod, nonpara, simclr, moco} and recently showed its great potential in conditional image synthesis \cite{park2020contrastive,contragan} by maintaining the correspondences between each generated image and its conditional input.
In this work, we further demonstrate how to
adapt it to \textit{diverse} conditional image synthesis by our newly proposed latent-augmented contrastive loss.

Specifically, conditioned on the same class code, the introduced latent-augmented contrastive loss forces the visual relations of the generated images to be correlated to the distances between their input latent codes, \textit{i.e.} ``positive'' images with close latent codes should be similar while ``negative'' images with distinct latent codes should be far away from each other in the feature space.
Note that the critical view (``positive'' or ``negative'') selection in our proposed framework is achieved via a novel latent augmentation scheme, \textit{i.e.} a positive code is sampled within a small hyper-sphere around the query code in the latent space while negative ones are sampled outside that hyper-sphere.
As a result, the problem of mode collapse can be alleviated to large extent by regularizing the generator on better understanding the structure of latent space.
Better performance on both learned distributions (Fig.~\ref{fig:toy}(d)) and generation results (Section~\ref{experiment}) indicate the effectiveness of our method.

Our contributions are summarized as three-fold:
\begin{itemize}
  \item We for the first time adapt contrastive learning to encourage diverse image synthesis.
  The proposed DivCo learning scheme can be readily integrated into existing conditional generative adversarial networks with marginal modifications.

  \item A novel latent-augmented contrastive loss is proposed to discriminate the latent representations of generated samples in a contrastive manner. The issue of mode collapse in cGANs has been much alleviated.

  \item Extensive experiments in different conditional generation tasks demonstrate that our proposed method helps existing frameworks improve the performance of diverse image synthesis without sacrificing the visual quality of the generated images.
\end{itemize}

\section{Related Work}
\noindent \textbf{Conditional generative adversarial network} was first presented in~\cite{cgan}. By feeding class labels into generator and discriminator simultaneously, it successfully generate images with respect to the given class labels. To further enhance the relation between input class labels and generated results, ACGAN~\cite{acgan} introduces an auxiliary classifier with cross entropy loss for categorizing both real and generated samples correctly. Similar scheme is also used in semi-supervised GANs~\cite{semigan, infogan, caglow}. After that, self-attention GAN and its follow-up BigGAN verify the effectiveness of conditional inputs on stabilizing and improving the training of GANs~\cite{spectralnorm, sagan, biggan}. Similarly, CocoGAN feeds into coordinate conditions into the generator~\cite{cocogan}. By changing the modality of conditions into texts, text-to-image generation realizes to synthesize images with respect to the given sentences~\cite{stackgan, stackgan2, attngan}. Nevertheless, image-to-image translation methods benefit from the development of cGANs most due to the large quantity of applications based on image processing. Pix2pix~\cite{pix2pix} and CycleGAN~\cite{cyclegan} translate different styles between two characteristic domains under paired and unpaired settings respectively. There are other applications of cGANs that condition on images such as super-resolution~\cite{srgan}, inpainting~\cite{inpainting-ca}, denoising~\cite{denoisinggan}, domain translation~\cite{cycada, Deng_2018_CVPR, stereogan} and so on~\cite{ganoverview, 9022578}. In this work, we endorse these conditional image synthesis tasks with the capability of generating more diverse samples with marginal modifications to their original frameworks.

\begin{figure*}[t]
    \centering
    \includegraphics[width=0.95\linewidth]{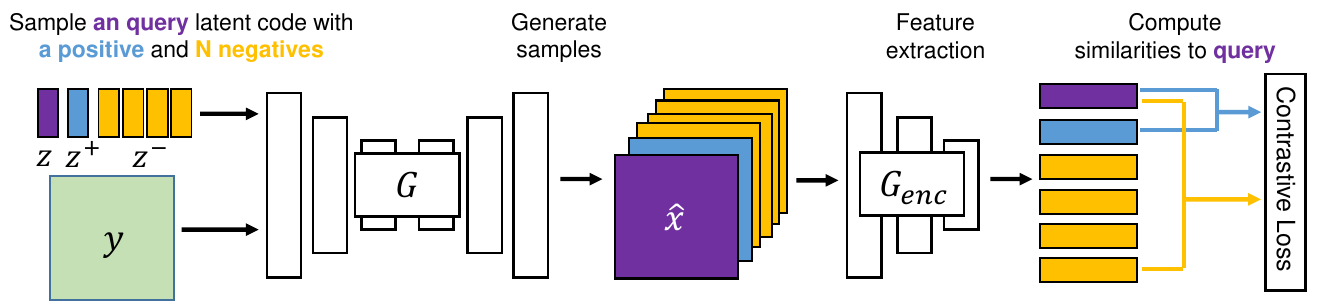}
    \caption{The proposed DivCo framework for image-to-image translation for achieving diverse conditional image synthesis. For a conditional image input $y$, we sample a query latent code $z$ with a positive code $z^{+}$ and $N$ negatives $z^{-}$ from a prior Gaussian distribution, which are fed into the generator $G$ to obtain the generated samples $\hat{x}$, $\hat{x}^{+}$, $\{\hat{x}^{-}_i\}_{i=1}^N$.
    The generated samples are passed into an auxiliary encoder $E$ to obtain their corresponding feature representations $f$, $f^{+}$, $\{f^{-}_i\}_{i=1}^N$ and are supervised by the contrastive loss.
    Here the encoder $E$ shares weights with the encoding layers of the generator $G_{enc}$. Note that we omit the original cGAN losses for better illustration. }
    \label{fig:framework}
\vspace{-1em}
\end{figure*}

\noindent \textbf{Contrastive learning}
was found effective in state-of-the-art self-supervised learning methods \cite{precod, nonpara, simclr, moco, tian2019contrastive}.
They generally cast the unsupervised visual representation learning as an instance discrimination task, where each unlabeled training sample is treated as a unique class.
A contrastive loss \cite{hadsell2006dimensionality,precod} is therefore introduced to make representations of different views from the same image closer and keep representations of different images away.
How to properly create the positive pairs becomes crucial in these methods, which can be different randomly augmented versions of a single image \cite{simclr}, a same image from various modalities \cite{tian2019contrastive}, and so on.

In generative modeling, some recent works investigate the use of contrastive loss for different objectives~\cite{park2020contrastive, TUNIT, contragan}. CUT~\cite{park2020contrastive} aimed to replace cycle-consistency loss with a patch-wise contrastive loss for improving the performance of unpaired image-to-image translation.
They considered positive pairs as images patches at the same location in two image domains and negative ones as patches from different locations within a single domain.
TUNIT~\cite{TUNIT} and ContraGAN~\cite{contragan} adopted contrastive losses for preserving the learned style (domain class) codes and categorical labels on the generated images respectively.
Note that all the above methods cannot be adopted for diverse image synthesis and cannot deal with mode collapse as well.
Our proposed DivCo is a totally different and novel contrastive learning strategy, compared to the existing methods.

\section{Method}

\begin{figure}[h]
    \centering
    \includegraphics[width=1\linewidth]{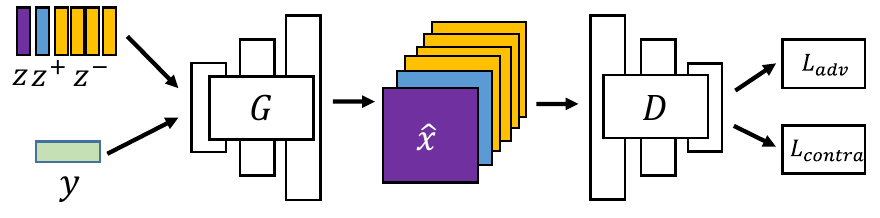}
    \caption{The proposed DivCo framework for class-conditioned image generation. Here $y$ represents conditional class label. Here the encoder $E$ shares weights with the encoding layers of discriminator $D$. }
    \label{fig:framework2}
\vspace{-1em}
\end{figure}

Given an image dataset $\mathcal{X}$ and its associated condition set $\mathcal{Y}$, we would like to learn a generator $G$ that synthesizes diverse images in the domain of $\mathcal{X}$ following the given conditions $\mathcal{Y}$.
% maps conditions $\mathcal{Y}$ to images $\mathcal{X}$ and synthesizes diverse images under certain conditions.
The generated images are expected to be realistic and diverse simultaneously.
% Such a generation process should meet the requirements of realism and diversity simultaneously.
Our proposed DivCo learning scheme can operate regardless of the exact forms of the condition dataset $\mathcal{Y}$,
% in various conditional generation tasks,
\textit{e.g.} prior images for paired or unpaired image-to-image translation tasks (Fig.~\ref{fig:framework}), class labels for class label-conditional generation tasks (Fig.~\ref{fig:framework2}), \textit{etc.}
The key innovation of our DivCo lies in the latent-augmented contrastive regularization term, which effectively avoids mode collapse and encourages more diverse conditional image synthesis. Note that this term can be seamlessly integrated into the training of any existing conditional generative adversarial network (cGANs) of different generation tasks without modifying their architectures.
% In this part, we would show how to integrate our proposed latent contrastive regularization loss into the training of conditional generative adversarial networks seamlessly and how our DivCo method works.

Formally, given an arbitrary condition $y\in\mathcal{Y}$ and a latent code $z\sim\mathcal{N}(0,1)$ sampled from a prior Gaussian distribution, the generator $G$ takes both $y$ and $z$ as inputs and yields an output image $\hat{x} = G(z,y)$.  Ideally, $\hat{x}$ should be determined by the condition $y$ and also be influenced by the latent code $z$.

\noindent \textbf{Adversarial loss.} The loss is adopted to ensure that $\hat{x}$ is vivid enough to be indistinguishable from real images $\mathcal{X}$ by a discriminator $D$.
% cGAN generally consists of a generator $G$ and a discriminator $D$. Given an arbitrary condition $y\in\mathcal{Y}$ from the condition dataset and a latent code $z\sim\mathcal{N}(0,1)$ sampled from a prior Gaussian distribution, the generator $G$ takes both $y$ and $z$ as inputs and yields an output image $\hat{x} = G(z,y)$.  $\hat{x}$ is determined by $y$ and meanwhile manipulated by $z$.
% , \textit{i.e.}, $\hat{x} = G(z,y)$.
% The aim of the generator is to synthesize such an image $\hat{x}$ that it is not only with respect to $y$ and $z$ but also vivid enough to be indistinguishable from real images. On the contrary, the discriminator $D$ learns to tell generated images from real ones correctly.
$G$ and $D$ play a minimax game and finally converge to a Nash equilibrium where the synthetic images cannot be distinguished from real images by $D$ any more. The adversarial loss is formulated as
%\begin{equation}
%  \label{eq:optim}
%  \begin{aligned}
%    \max_{D} \min_{G} \mathcal{L}(G,D),
%  \end{aligned}
%\end{equation}
%where $\mathcal{L}=\mathcal{L}_{adv}$ is the adversarial loss between $G$ and $D$:
\begin{equation}
\label{eq:advloss}
\begin{aligned}
  \mathcal{L}_\text{adv}  &(G, D) = \mathbb{E}_{x\sim\mathcal{X}, y\sim\mathcal{Y}} \left[ \log{D(x,y)} \right] \\
  &+  \mathbb{E}_{y\sim\mathcal{Y}, z \sim \mathcal{N}(0,1)} \left[ \log{ (1 - D(G(z,y), y) } \right].
\end{aligned}
\end{equation}
% Such adversarial loss shows great potential in synthesizing realistic images and its output can be conditioned on the input well, but it usually gets stuck in lacking of diversity due to the issue of mode collapse.
However, it can only encourage the realism rather than the diversity of the generated images.

% Such adversarial loss can promote the realism of generated images but fails on encouraging the diversity of generated images.

\noindent \textbf{Diverse image synthesis via contrastive learning}.
% We aim to alleviate this issue in cGANs of neural networks in many different conditional generation tasks.
We aim to properly regularize cGANs on diverse conditional image synthesis with properly designed learning objectives instead of modifying the original architectures to adapt to different conditional generation tasks.

Intuitively, to avoid mode collapse, we should strengthen the influence of $z$ to the generated image $\hat{x}$ while preserving the fitness of the image $\hat{x}$ to the conditional input $y$.
%a dominant position for latent codes $z$ on the output images $\hat{x}$ while preserving the conditional information $y$ and avoiding sacrificing visual quality.
In other words, the variations of $\hat{x}$ should be intently correlated to the manipulation of $z$.
% the mutual information between latent codes and outputs should be maximized.
To achieve this goal, we propose to adopt contrastive learning, where the goal is to associate a query image and its ``positive'' sample, in contrast to other ``negatives'' samples at the same time for supervising the image generation process.
%A softmax-like function can be adopted to perform a $(N+1)$-way classification problem for supervision.
%\begin{equation}
%\label{eq:cls}
%\begin{aligned}
%  & \ell (q,p,k) =\\
%  &-\log \left[ \frac{\exp{(\langle q, p \rangle %/\tau)}}{\exp{(\langle q, p \rangle / \tau)} + \sum_{i=1}^{N} %{\exp{(\langle q, k_i \rangle / \tau)}}} \right],
%\end{aligned}
%\end{equation}
%where $\langle\cdot,\cdot\rangle$ denotes the inner product between two normalized feature vectors to measure their similarity and $\tau$ is a temperature hyper-parameter for scaling the similarity.
% The basic concept and relation of these three terms are exploited in this work but essentially different from that of visual contrastive representation learning~\cite{moco}, which would be discussed later.
The most critical design in contrastive learning is how to select the ``positive'' and ``negative'' pairs.

Specifically, given a certain condition $y$, we wish images generated from close (``positive'') latent codes to be visually similar and images from far-away (``negative'') latent codes to be visually dissimilar.
For a query generated image $\hat{x}=G(z,y)$ generated from a latent code $z$,
we denote its ``positive'' code as $z^+$ and the corresponding generated image can be represented as $\hat{x}^+=G(z^+,y)$.
Similarly, the generated images for a series of ``negative'' latent code $\{z_i^-\}_{i=1}^N$ can be represented as $\hat{x}^-_i=G(z^-_i,y)$ for $i=1,\dots,N$.
Taking advantages of an auxiliary encoder $E$ (will be discussed), we extract feature representations for generated images, \textit{i.e.}, $f=E(\hat{x})$, $f^+=E(\hat{x}^+)$, $f^-_i=E(\hat{x}^-_i)$.
These representations are expected to be discriminative and can be adopted in contrastive learning to supervise the image generator $G$.

% \noindent \textbf{Latent contrastive loss}.
A novel regularization term, namely {\it latent-augmented contrastive loss}, is proposed to regularize the generated images' feature representations $f, f^+, f^-$ by bringing $f$, $f^+$ closer and spreading $f$, $f^-$ away. The regularization term is formulated in the form of contrastive loss as
\begin{align}
  & {\cal L}_\text{contra}(G) = \label{eq:cls} \mathbb{E}_{y\sim\mathcal{Y},z\sim\mathcal{N}(0,1)} \\
  & \left[ -\log  \frac{\exp{(\langle f, f^+ \rangle /\tau)}}{\exp{(\langle f, f^+ \rangle / \tau)} + \sum_{i=1}^{N} {\exp{(\langle f, f^-_i \rangle / \tau)}}} \right], \nonumber
\end{align}
%\begin{equation}
%\label{eq:contraloss}
%\begin{aligned}
%  & \mathcal{L}_\text{contrast} (G,E) = %\mathbb{E}_{y\sim\mathcal{Y},z\sim\mathcal{N}(0,1)}[\ell(f,f^+,f^-)],
%\end{aligned}
%\end{equation}
where $\langle\cdot,\cdot\rangle$ denotes the inner product between two normalized feature vectors to measure their similarity, and $\tau$ is a temperature hyper-parameter for scaling the similarity.
% The basic concept and relation of these three terms are exploited in this work but essentially different from that of visual contrastive representation learning~\cite{moco}, which would be discussed later.
For each query generated image $\hat{x}$, we generate a positive sample $\hat{x}^+$ and multiple negative samples $\{\hat{x}_i^-\}_{i=1}^N$ in training.

We create positive and negative samples, $\hat{x}^+$ and $\{\hat{x}^-_i\}$, for the latent-augmented contrastive loss by manipulating the input latent codes, $\hat{z}^+$ and $\{\hat{z}^-_i\}$. We observe that the strategy for creating $\hat{z}^+$ and $\{\hat{z}^-_i\}$ has critical impact to the contrastive learning (similar observations can be found in the self-supervised representation learning tasks \cite{simclr, moco} and image synthesis \cite{park2020contrastive, TUNIT}).
We propose a \textit{latent-augmentation} strategy to generate positive and negative samples for contrastive learning on-the-fly.
Note that such a newly proposed augmentation strategy is conducted on the latent space rather than the image manifold as existing unsupervised learning methods \cite{simclr, moco}, to adapt to our generation tasks.
% so as to make the generator models an unbiased distribution in this work.
% \noindent \textbf{Latent Augmentation}. Similar to contrastive representation learning tasks, we must choose well-defined ``positive'' and ``negative'' samples as well. In contrastive representation learning tasks, they choose a positive image by augmenting the query image by rotation, flipping, random crop and so on. The feature representation of this augmented image should be similar to that of query image~\cite{moco}. On the contrary, we should define augmenting strategy on the latent space instead of image manifold so as to make the generator models an unbiased distribution in this work.
Specifically, given a query latent code $z\sim \mathcal{N}(0,1)$,
% After sampling a query latent code $z$ randomly from a prior Gaussian distribution,
we define a small hyper-sphere with radius $R$ centered at the query code $z$.
Consequently, a positive latent code $z^+$ can be randomly sampled as a vector within the sphere $z^+ = z + \delta$, where $\delta$ is a randomly sampled vector from a uniform distribution $\delta \sim \mathcal{U}[-R, R]$,
% We then randomly sample a latent code from inner sphere $|z|<\delta$ and regard this latent code as a positive.
while the negative latent codes can be sampled as random vectors outside the sphere within the latent space, \textit{i.e.}, $z^-_i \sim \{\left. z^-_i \right| z^-_i \sim {\cal N}(0,1) \cap |z^-_i - z| \succ R\}$ for $i=1, \dots, N$ where $\succ$ is element-wise greater-than operator.
%And those latent codes sampled from outer sphere $|z^--z|>\delta$ are automatically referred to as negatives $\{z^-\}$.

% \noindent \textbf{Auxiliary encoder}. Due to the huge computational complexity of calculating the similarity in image manifold, we use an auxiliary encoder to extract the feature representation of those generated images. The auxiliary encoder should possess a great capability of producing high-quality features for representing high-dimensional images. Thus we could calculate the similarity score between the generated images by replacing them with their representations.

The generator $G$ itself in cGANs is able to encode discriminative feature representations as discussed in \cite{park2020contrastive}. Therefore, we directly adopt the encoding layers of the generator as our auxiliary encoder $E$, as shown in Fig.~\ref{fig:framework}.
%without introducing new architectures for different cGANs.
Note that since class-conditioned image generation tasks do not require encoding layers in the generator, we use the encoding layers of the conditional discriminator instead, which are also able to encode discriminative features, as shown in Fig.~\ref{fig:framework2}.
By exploiting the off-the-shelf generators or discriminators for extracting feature representations in the contrastive learning,
we could avoid modifying the original architectures of different conditional generation methods, showing the plug-and-play functionality of our method.

\noindent \textbf{Overall training objective}.
Our proposed latent-augmented contrastive loss can be readily integrated into existing cGANs for different conditional generation tasks, \textit{i.e.} simply adding our proposed latent contrastive loss with their original training objectives. The overall training objective can be formulated as
% we just adopt their original optimization objectives and add our proposed latent contrastive loss to it, forming our full objective which can be formulated as:
\begin{equation}
  \label{eq:final}
  \begin{aligned}
    \max_{D} \min_{G}
     \mathcal{L}_\text{adv}(G,D)
    + \lambda_\text{contra}\mathcal{L}_\text{contra}(G) + \lambda_\text{opt} \mathcal{L}_\text{opt}(G) ,
  \end{aligned}
\end{equation}
where $\lambda_\text{opt}$ and $\lambda_\text{contra}$ are hyper-parameters balancing different losses in different generation tasks. $\mathcal{L}_\text{opt}$ represents an optional regularization term of different conditional generation tasks.

$\mathcal{L}_\text{opt}$ is not required for class label-conditional generation tasks.
% For image-to-image translation tasks, however, it is regarded that adding an extra pixel-level regularization loss may be the best option.
For paired image-to-image translation, the $\mathcal{L}_\text{opt}$ regularization term can be formed as a pix-to-pix reconstruction loss
\begin{equation}
\label{eq:sup-i2i}
\begin{aligned}
  & \mathcal{L}_\text{opt} (G) = \mathbb{E}_{x\sim\mathcal{X}, y\sim\mathcal{Y}, z \sim \mathcal{N}(0,1)} \left[ \left\|G(z,y) - x \right\|_1 \right].
\end{aligned}
\end{equation}
For unpaired image-to-image translation, $\mathcal{L}_\text{opt}$ can be formed as the cyclic reconstruction loss,
% \begin{equation}
\begin{align}
\label{eq:uns-i2i}
  \mathcal{L}_\text{opt} (G) = &\mathbb{E}_{x\sim\mathcal{X}, y\sim\mathcal{Y}, z \sim \mathcal{N}(0,1)} \Big[ \left\|F(z, G(z,y)) - y \right\|_1  \nonumber\\
                             & +  \left\|G(z, F(z,x)) - x \right\|_1  \Big] ,
\end{align}
% \end{equation}
where $F$ is an auxiliary generator which learns the inverse mapping of $G$ from $\mathcal{X}$ to $\mathcal{Y}$.

\noindent \textbf{Discussion}.
In this work, we for the first time investigate adopting contrastive learning for diverse conditional image synthesis. We propose a novel latent augmentation strategy for creating ``positive'' and ``negative'' samples in the latent space, while all the previous contrastive losses \cite{moco, simclr, park2020contrastive, TUNIT} focus on image-based augmentation strategies that cannot be used for the diverse image synthesis task.

\begin{figure*}[t]
    \centering
    \includegraphics[width=1\linewidth]{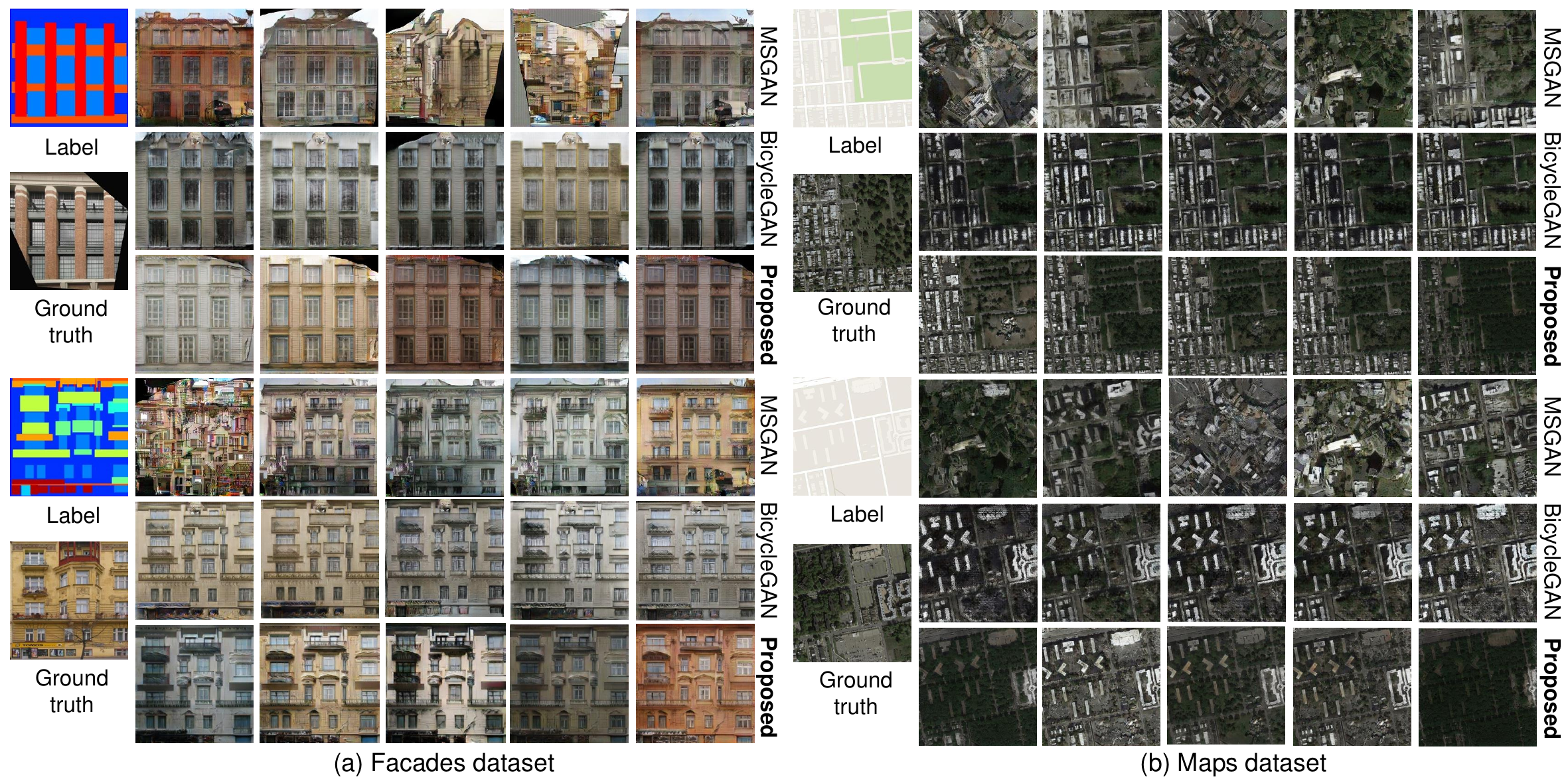}
    \caption{\textbf{Diversity comparison} in the paired image-to-image translation task on Facades and Maps datasets. MSGAN produces diverse images but sacrifices the visual quality and even sometimes cannot condition on the input labels. BicycleGAN produces more realistic images but its results lack diversity. Our proposed method synthesizes both realistic and diverse images.}
    \label{fig:paired}
\vspace{-1em}
\end{figure*}

In diverse image synthesis, previous regularization terms~\cite{bicyclegan, DRIT, msgan} on encouraging diversity synthesis failed on jointly modeling positive and negative relations between the generated images, which are sub-optimal solutions.
In addition, unlike mode seeking loss that actually constrains the generator to an opposite direction to pixel-level reconstruction losses, which easily breaks down (see Fig.~\ref{fig:paired}), our latent-augmented contrastive loss %helps learn more discriminative feature representations and it
can be combined with original cGAN losses in a harmonious manner, leading to better performance in diverse image synthesis.
%The generator trained by our proposed DivCo could model unbiased distributions (see Fig.\ref{fig:toy}) and thus synthesizes more diverse images (see Section~\ref{experiment}).

\section{Experiment}
\label{experiment}

\begin{table}\footnotesize
\begin{center}
\caption{\label{table:paired}Quantitative results of paired image-to-image translation on Facades and Maps dataset. }
\begin{spacing}{1.20}
\begin{tabular}{l|c|c|c}
\hline
\cline{1-4}
      \multicolumn{4}{c}{Facades}  \\
\cline{1-4}
%{\backslashbox{Metric}{\kern-2em Method}} & BicycleGAN & MSGAN & Proposed  \\
Metric    & BicycleGAN~\cite{bicyclegan} & MSGAN~\cite{msgan} & Proposed  \\
\hline
NDB   $\downarrow$  & $8.00\pm 1.00$ & $7.60\pm 0.94$ & $\textbf{7.30}\pm 0.66$ \\
\hline
JSD   $\downarrow$  & $0.0427\pm 0.0045$ & $0.0392\pm 0.0088$ & $\textbf{0.0374}\pm 0.0098$ \\
\hline
LPIPS  $\uparrow$  & $0.2355\pm 0.0067$ & $\textbf{0.2677} \pm 0.0259$ & $0.2531 \pm 0.0108$ \\
\hline
FID  $\downarrow$  & $80.34\pm 1.46$ & $78.32\pm 3.32$ & $\textbf{75.96}\pm 1.25$ \\
\hline
\hline
      \multicolumn{4}{c}{Maps}  \\
\cline{1-4}
%{\backslashbox{Metric}{\kern-2em Method}} & BicycleGAN & MSGAN & Proposed  \\
Metric    & BicycleGAN~\cite{bicyclegan} & MSGAN~\cite{msgan} & Proposed  \\
\hline
NDB    $\downarrow$ & $8.90\pm 1.86$ & $8.60\pm 1.98$ & $\textbf{7.80}\pm 2.35$ \\
\hline
JSD    $\downarrow$ & $0.0956\pm 0.0030$ & $0.1066\pm 0.0023$ & $\textbf{0.0806}\pm 0.0021$ \\
\hline
LPIPS  $\uparrow$ & $0.2473\pm 0.0226$ & $0.2306 \pm 0.0287$ & $\textbf{0.2538} \pm 0.0382$ \\
\hline
FID   $\downarrow$ & $130.74\pm 2.63$ & $169.97\pm 8.62$ & $\textbf{109.54}\pm 4.41$ \\
\hline
\cline{1-4}
\end{tabular}
\end{spacing}
\end{center}
\vspace{-3.5em}
\end{table}

\begin{figure*}[t]
    \centering
    \includegraphics[width=0.995\linewidth]{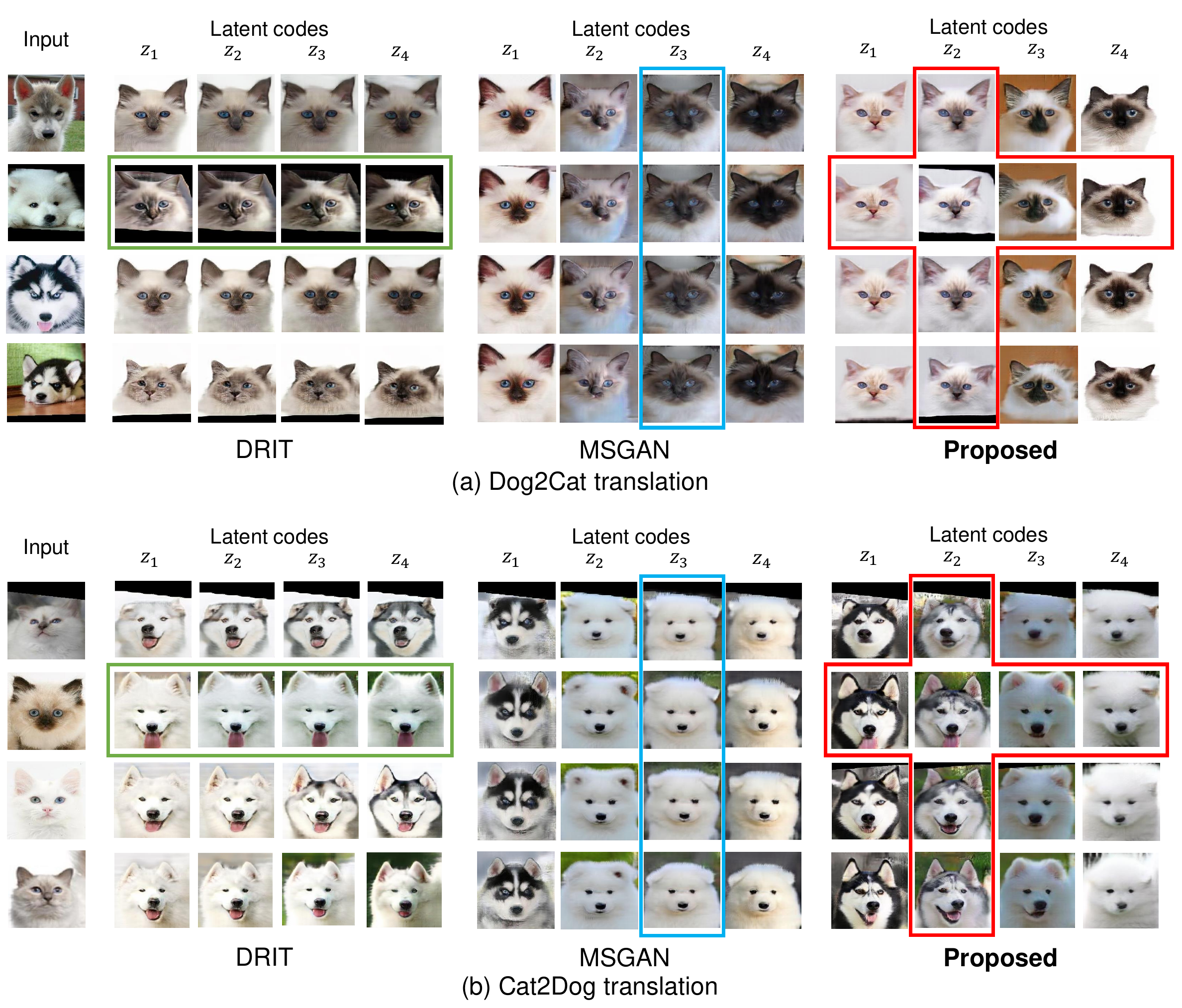}
    \caption{\textbf{Diversity Comparison} in the unpaired image-to-image translation task on shape-variant dataset Dog$\Leftrightarrow$Cat. We manipulate the latent codes from $z_1$ to $z_4$ where $z_1$ is an all-negative-one vector, $z_4$ is an all-one vector, and $z_2$ and $z_3$ are the interpolation of $z_1$ and $z_4$ with equal intervals. DRIT produces images without shape variation (mode collapse). MSGAN produces images that are badly conditioned on the input image. Our proposed method synthesizes diverse images by either changing the input condition or manipulating the input latent code.}
    \label{fig:dog2cat}
\vspace{-0.5em}
\end{figure*}

\subsection{Implementation details}

We evaluate our proposed DivCo framework with extensive qualitative and quantitative experiments. We integrate the proposed latent-augmented contrastive loss in three conditional generation tasks, including paired image-to-image translation, unpaired image-to-image translation and class label-conditioned generation. We would maintain their original network architectures and supervision signals. We only add our latent-augmented contrastive loss to their optimization objectives and adjust the relative weights between the loss terms. Empirically, the number of negative samples $N$ is set to $10$, the balancing weight for latent-augmented contrastive loss $\lambda_\text{contra}$ is set to $1$ and the temperature $\tau$ is set to $1$ throughout our experiments if not particularly indicated. For the radius $R$ which depends on the dimension of latent code to some extent, is set to $0.01$ in image-to-image translation tasks and set to $0.001$ in class-conditioned image generation task.

\noindent \textbf{Baselines.} For paired image-to-image translation tasks, we choose BicyleGAN~\cite{bicyclegan} as our baseline. For unpaired image-to-image translation, DRIT~\cite{DRIT} is chosen as our baseline. Both of them utilize a latent regression loss for building correlation between latent codes and generated images. %It also uses latent regression loss to help making generated images various by manipulating latent codes.
We remove their original latent regression loss and then add mode seeking loss~\cite{msgan} and our proposed latent-augmented contrastive loss separately for fair comparison.
As for class-conditioned generation, we take DCGAN~\cite{dcgan} as our baseline following~\cite{msgan}. Since its generator do not require encoding layers, we take all but the topmost layers of the discriminator as our auxiliary encoder for calculating latent-augmented contrastive loss.

\noindent \textbf{Datasets.} We use Facades~\cite{facades} and Maps~\cite{pix2pix} datasets for evaluating paired image-to-image translation performance.
For unpaired image-to-image translation, we take Yosemite~\cite{cyclegan} and Dog2Cat~\cite{DRIT} datasets for evaluation. CIFAR10~\cite{Krizhevsky09} is utilized for evaluating class-conditioned generation results.

\noindent \textbf{Evaluation metrics}. We evaluate the performance of the compared methods via calculating following evaluation metrics. {\it NDB} and {\it JSD} are two bin-based evaluation metrics for measuring the similarities between generated distributions and real distributions~\cite{ndb}. Lower value means higher similarity. {\it Fr\'echet Inception Distance (FID)}, proposed in~\cite{fid}, is used to measure the quality and diversity of a set of generated images compared to real images by feeding them into an Inception network~\cite{inception}. Lower value means higher visual quality and higher diversity. {\it LPIPS} is used to measure the diversity of an image set by computing the average feature distance of all pairs of images~\cite{lpips}. Higher value means higher inter-sample diversity.

\begin{table*}[t]
\small
\begin{center}
\caption{\label{table:unpaired}Quantitative results of unpaired image-to-image translation on Cat$\Leftrightarrow$Dog and Yosemite (Winter$\Leftrightarrow$Summer) dataset.}
\vspace{0em}
\begin{spacing}{1.19}
\begin{tabular}{l|c|c|c|c|c|c}
\hline
\cline{1-7}
Dataset  &   \multicolumn{3}{c|}{Dog2Cat}    &    \multicolumn{3}{c}{Cat2Dog}  \\
\cline{1-7}
%{\backslashbox{Metric}{\kern-2em Method}} & BicycleGAN & MSGAN & Proposed  \\
       & DRIT \cite{DRIT} & MSGAN \cite{msgan} & Proposed    & DRIT \cite{DRIT} & MSGAN \cite{msgan} & Proposed \\
\hline
NDB   $\downarrow$  & $5.20\pm 1.72$ & $4.20\pm 1.32$ & $\textbf{3.60}\pm 0.80$ & $7.7\pm 1.00$ & $6.30 \pm 1.13$ & $\textbf{5.90}\pm 1.04$ \\
\hline
JSD   $\downarrow$  & $0.0362 \pm 0.0035$ & $0.0285\pm 0.0042$ & $\textbf{0.0238}\pm 0.0039$ & $0.0440\pm 0.0036$ & $0.0381\pm 0.0043$ & $\textbf{0.0345}\pm 0.0063$ \\
\hline
LPIPS  $\uparrow$   & $0.1932 \pm 0.0059$ & $0.2057\pm 0.0052$ & $\textbf{0.2188}\pm 0.0045$ & $0.2544 \pm 0.0145$ & $0.2615\pm 0.0156$ & $\textbf{0.2686}\pm 0.0214$\\
\hline
FID  $\downarrow$  & $30.40 \pm 1.82$ & $27.46\pm 1.89$ & $\textbf{24.83}\pm 2.40$ & $21.54\pm 2.34$ & $20.78\pm 1.09$ & $\textbf{19.25}\pm 2.14$ \\
\hline
\hline
Dataset  &   \multicolumn{3}{c|}{Winter2Summer}    &    \multicolumn{3}{c}{Summer2Winter}  \\
\cline{1-7}
%{\backslashbox{Metric}{\kern-2em Method}} & BicycleGAN & MSGAN & Proposed  \\
       & DRIT \cite{DRIT} & MSGAN \cite{msgan} & Proposed & DRIT \cite{DRIT} & MSGAN \cite{msgan} & Proposed \\
\hline
NDB   $\downarrow$  & $5.80\pm 2.34$ & $5.00\pm 1.56$ & $\textbf{4.90}\pm 1.30$ & $6.30\pm 1.96$ & $5.70 \pm 1.01$ & $\textbf{5.40}\pm 0.91$ \\
\hline
JSD   $\downarrow$  & $0.0584 \pm 0.0079$ & $0.0401\pm 0.0097$ & $\textbf{0.0338}\pm 0.0059$ & $0.0625\pm 0.0052$ & $0.0595\pm 0.0080$ & $\textbf{0.0573}\pm 0.0057$ \\
\hline
LPIPS  $\uparrow$  &  $0.1801 \pm 0.0021$ & $0.1816\pm 0.0043$ & $\textbf{0.1869}\pm 0.0053$ & $0.1721\pm 0.0072$ & $0.1795\pm 0.0080$ & $\textbf{0.1926}\pm 0.0093$\\
\hline
FID  $\downarrow$  & $47.58 \pm 2.12$ & $46.37\pm 1.90$ & $\textbf{45.79}\pm 2.45$ & $56.02\pm 2.10$ & $53.32\pm 1.67$ & $\textbf{51.57}\pm 1.92$ \\
\hline
\cline{1-7}
\end{tabular}
\end{spacing}
\end{center}
\vspace{-3.5em}
\end{table*}

\subsection{Paired image-to-image translation}

%\begin{figure}[t]
%    \centering
%    \includegraphics[width=1\linewidth]{exp_facades.pdf}
%    \caption{Qualitative results on Facades dataset. }
%    \label{fig:facades}
%\end{figure}
%\begin{figure}[t]
%    \centering
%    \includegraphics[width=1\linewidth]{exp_maps.pdf}
%    \caption{Qualitative results on Maps dataset. }
%    \label{fig:maps}
%\end{figure}

We take BicycleGAN~\cite{bicyclegan} as our baseline and separately replace the latent regression loss with mode seeking loss~\cite{msgan} and our proposed latent-augmented contrastive loss for fair comparison. We show our generation results on Facades dataset in Fig.~\ref{fig:paired}(a). The generation results of MSGAN are diverse enough for each input condition, but the visual quality drops drastically. On the contrary, the generation results of BicycleGAN and our method looks more realistic. In addition, our method's results show better generation diversity. Similar comparisons could be found on the Maps dataset, as shown in Fig.~\ref{fig:paired}(b). There is another interesting finding that MSGAN produces similar patterns on the $3$-th and $4$-th columns regardless of the input conditional label maps, which is consistent to the toy demonstration in Fig.~\ref{fig:toy}.

The quantitative results on Facades and Maps datasets are recorded in Table~\ref{table:paired}. As one can see, our method performs best in terms of all the evaluation metrics except for LPIPS on Facades. The reason is that LPIPS only measure the diversity score of a set of images by their average feature distance while the realism is not taken into consideration.

\begin{table*}
\begin{center}
\caption{\label{table:class-ndb}NDB and JSD of the generated images by different compared methods on CIFAR10.}
\vspace{-1em}
\begin{spacing}{1.15}
\begin{tabular}{c|c|c|c|c|c|c}
  \hline
  \cline{1-7}
  Metric               & Method & airplance & automobile & bird & cat  & deer   \\
  \cline{3-7}
  \multirow{3}{*}{NDB$\downarrow$} & DCGAN \cite{dcgan} & $17.90 \pm 1.56$ & $18.40 \pm 1.82$ & $13.80 \pm 2.36$ & $14.50 \pm 1.90$ & $18.90 \pm 2.18$  \\
                       & MSGAN \cite{msgan}            & $17.40 \pm 2.80$ & $17.10 \pm 2.07$ & $14.10 \pm 2.62$ & $16.50 \pm 2.85$ & $20.70 \pm 2.45$  \\
                       & Proposed & $\textbf{16.30} \pm 2.75$ & $\textbf{16.90} \pm 1.92$ & $\textbf{12.80} \pm 2.14$ & $\textbf{10.70} \pm 2.49$ & $\textbf{14.60} \pm 1.96$  \\
  \hline
  \multirow{3}{*}{JSD$\downarrow$} & DCGAN \cite{dcgan} & $0.0098\pm 0.0001$ & $0.0125\pm 0.0001$ & $0.0082\pm 0.0001$ & $0.0086\pm 0.0001$ & $0.0069\pm 0.0001$  \\
                       & MSGAN \cite{msgan} & $0.0095\pm 0.0001$ & $0.0121\pm 0.0001$ & $0.0083\pm 0.0002$ & $0.0154\pm 0.0001$ & $0.0072\pm 0.0001$  \\
                       & Proposed & $\textbf{0.0091}\pm 0.0001$ & $\textbf{0.0119}\pm 0.0001$ & $\textbf{0.0076}\pm 0.0001$ & $\textbf{0.0059}\pm 0.0002$ & $\textbf{0.0058}\pm 0.0001$  \\
  \hline
                       &       & dog & frog & horse & ship & truck   \\
  \cline{3-7}
  \multirow{3}{*}{NDB$\downarrow$} & DCGAN \cite{dcgan} & $14.50 \pm 1.83$ & $24.00 \pm 1.55$ & $24.60 \pm 2.36$ & $18.50 \pm 2.51$ & $15.90 \pm 1.33$  \\
                       & MSGAN \cite{msgan}            & $11.20 \pm 1.16$ & $22.30 \pm 2.90$ & $22.90 \pm 1.83$ & $16.40 \pm 3.36$ & $15.00 \pm 2.69$  \\
                       & Proposed & $\textbf{11.00} \pm 2.24$ & $\textbf{20.20} \pm 2.22$ & $\textbf{20.90} \pm 2.76$ & $\textbf{16.30} \pm 2.29$ & $\textbf{14.60} \pm 2.28$  \\
  \hline
  \multirow{3}{*}{JSD$\downarrow$} & DCGAN \cite{dcgan} & $0.0088\pm 0.0001$ & $0.0150\pm 0.0001$ & $0.0171\pm 0.0001$ & $0.0112\pm 0.0001$ & $0.0088\pm 0.0001$  \\
                       & MSGAN \cite{msgan} & $0.0072\pm 0.0003$ & $0.0134\pm 0.0001$ & $0.0154\pm 0.0001$ & $0.0096\pm 0.0001$ & $0.0084\pm 0.0001$  \\
                       & Proposed & $\textbf{0.0058}\pm 0.0002$ & $\textbf{0.0124}\pm 0.0001$ & $\textbf{0.0126}\pm 0.0001$ & $\textbf{0.0085}\pm 0.0001$ & $\textbf{0.0078}\pm 0.0002$  \\
  \hline
  \cline{1-7}
\end{tabular}
\end{spacing}
\end{center}
\vspace{-3em}
\end{table*}

\subsection{Unpaired image-to-image translation}

\begin{figure}[t]
    \centering
    \includegraphics[width=1\linewidth]{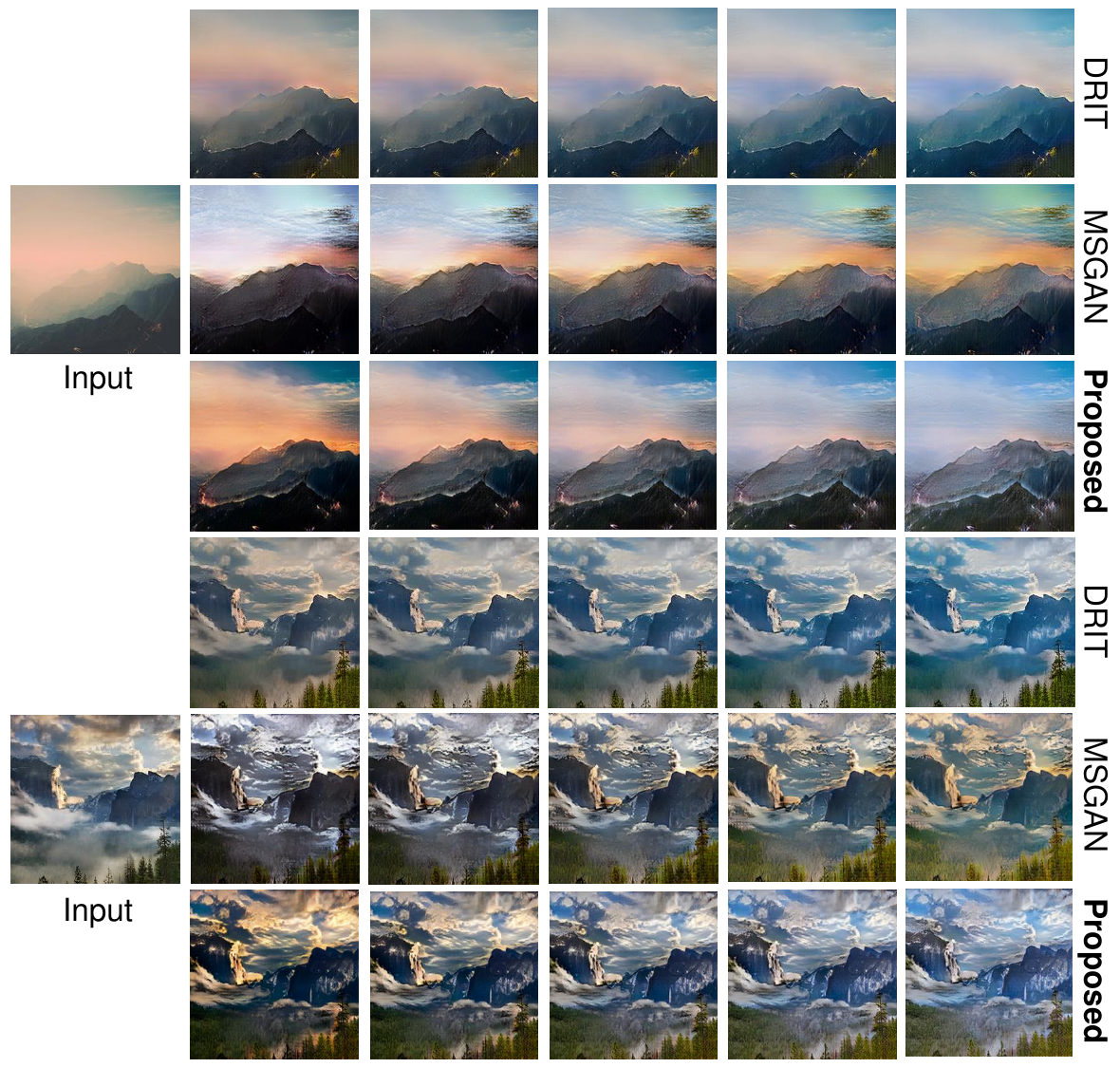}
    \caption{\textbf{Diversity comparison} in the unpaired image-to-image translation task on shape-invariant dataset Winter2Summer. Only the generation results from our proposed method shows the variation from sunshine appearing to totally disappearing, compared to two other methods.}
    \label{fig:w2s}
\vspace{-1.5em}
\end{figure}

In unpaired image-to-image translation, we choose DRIT~\cite{DRIT} as our baseline and replace its original latent regression loss with mode seeking loss and our latent contrastive loss for encouraging the diversity of the generation results. We choose a shape-variant dataset Cat$\leftrightarrow$Dog and a shape-invariant dataset Winter$\leftrightarrow$Summer for comparison.

The synthesized images on Cat$\leftrightarrow$Dog dataset are shown in Fig.~\ref{fig:dog2cat}. We highlight the advantages of our approach's generation results over other methods by the rectangles in different colors. For the generation results from DRIT, the green rectangle shows its incapability of synthesizing diverse images according to various latent codes. For the results from MSGAN, the blue rectangle highlights its incapability of synthesizing diverse images with  different conditional inputs, which could also be observed in Fig.~\ref{fig:toy} and Fig.~\ref{fig:paired}(b). The synthesized images from our proposed method show much higher diversity than those of previous works, as highlighted in the red cross. Our generated results also fit the different input conditions much better than MSGAN.
More diverse generation results can also be observed in Winter2Summer dataset in Fig.~\ref{fig:w2s}. %Only the generation results from our proposed method shows the variance of sunshine from appearing to totally disappearing, compared to the generation results from DRIT and MSGAN.

The quantitative results of different approaches on the two datasets are summarized in Table~\ref{table:unpaired}. As we see, the proposed method performs best in all evaluation metrics under different conditions compared to two other state-of-the-art diverse image synthesis methods, which demonstrate the effectiveness of our method on this task.

\subsection{Class-conditioned image generation}

We choose DCGAN~\cite{dcgan} as our baseline for comparing the performance on class-conditioned image generation. Since there is no encoder-like architecture in the generator of DCGAN and the dimension of intermediate discriminator feature does not match that of latent code, the latent regression loss is not applicable. Table~\ref{table:class-ndb} shows the quantitative results of NDB and JSD for each class. For calculating NDB and JSD, all training samples are first clustered into $50$ groups, and each generated sample is then assigned to a nearest group. The NDB score and JSD are then calculated according to the group proportions. We also calculate FID for all generated images, as shown in Table~\ref{table:class-fid}. Our method performs best in terms of all these evaluation metrics.

\begin{table}
\begin{center}
\caption{\label{table:class-fid}FID of the generated images by different compared methods on CIFAR10 dataset.}
\begin{spacing}{1.19}
\begin{tabular}{c|c|c|c}
\hline
\cline{1-4}
     Method  & DCGAN \cite{dcgan} & MSGAN \cite{msgan} & Proposed \\
\hline
     FID $\downarrow$ & $28.34\pm 0.12$ & $27.09\pm 0.14$ & $\textbf{25.21}\pm 0.20$ \\
\hline
\cline{1-4}
\end{tabular}
\end{spacing}
\end{center}
\vspace{-3em}
\end{table}

\subsection{Sensitivity Analysis}

\begin{figure}[t]
    \centering
    \includegraphics[width=1\linewidth]{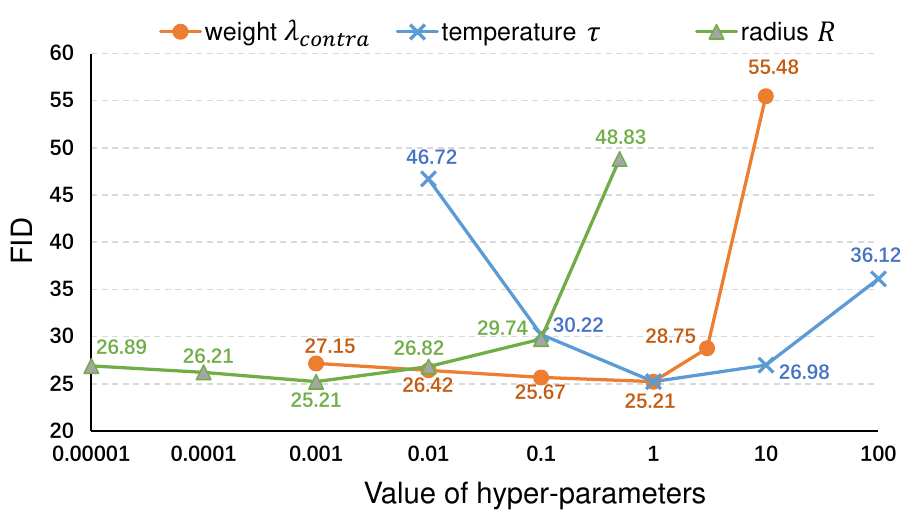}
    \caption{Sensitivity analysis for the relative weight of latent-augmented loss $\lambda_\text{contra}$, temperature for scaling similarity $\tau$, and the radius of hyper-sphere $R$ under class-conditioned image generation task. }
    \label{fig:sen}
\vspace{-1em}
\end{figure}

We perform sensitivity analysis for $3$ hyper-parameters in our framework: weight of latent-augmented loss $\lambda_\text{contra}$, temperature for scaling similarity $\tau$, and the radius of hyper-sphere $R$ in the class-conditioned image generation task. We empirically set $\lambda_\text{contra}=1$, $\tau=1$, and $R=0.001$. We take FID as our metric for balancing both realism and diversity. The FID results are plotted in Fig.~\ref{fig:sen}.

As we can see, when the weight $\lambda_\text{contra}$ decreases, the performance drops slightly and gradually approach the FID value of original DCGAN (see Table~\ref{table:class-fid}). When $\lambda_\text{contra}$ is greater than $3$, the performance drops drastically as $\lambda_\text{contra}$ increases, as ${\cal L}_{\rm contra}$ overwhelms the original cGAN loss, leading to images of lower realism.
The larger the radius of hyper-sphere $R$ is, the larger the distance between query and positive latent codes tends to be, leading to a more biased definition of ``positive'' and ``negative''. However, when $R$ becomes smaller and gradually approaches $0$, the effect of modeling positive relations between images tends to be weaker and make the proposed contrastive loss performs similarly to the mode seeking loss in MSGAN.
As for the temperature $\tau$, we find that only a proper value leads to the best performance. Either higher or lower values would make the latent-augmented contrastive loss generate inferior results.
%Higher temperature makes the optimization objective hard for learning discriminative feature representations. Lower temperature makes similarity scores higher, leading to it harder to find the optimal solution.

\section{Conclusion}
In this paper we propose a novel framework DivCo for dealing with mode collapse issue in conditional generative adversarial networks. We adapt contrastive learning to diverse image synthesis and propose a novel latent-augmented contrastive loss. The proposed method could be readily integrated into existing GAN-based conditional generation approaches by integrating the latent-augmented contrastive regularization term to their original optimization objective. Experiments on multiple settings with various datasets verify the superiority of our method to other state-of-the-arts, both qualitatively and quantitatively.

\noindent \textbf{Acknowledgement}. This work is supported in part by Centre for Perceptual and Interactive Intelligence Limited, in part by the General Research Fund through the Research Grants Council of Hong Kong under Grants (Nos. 14208417, 14207319, 14202217, 14203118, 14208619), in part by Research Impact Fund Grant No. R5001-18, in part by CUHK Strategic Fund.

{\small
\bibliographystyle{ieee_fullname}
%\bibliography{egbib}

}

\end{document}